\documentclass{article}

    \PassOptionsToPackage{numbers, compress}{natbib}

\usepackage[main, final]{neurips_2026}

\usepackage[utf8]{inputenc} 
\usepackage[T1]{fontenc}    
\usepackage{hyperref}       
\usepackage{url}            
\usepackage{booktabs}       
\usepackage{amsfonts}       
\usepackage{nicefrac}       
\usepackage{microtype}      
\usepackage{xcolor}         
\usepackage{amsmath}
\usepackage{graphicx}
\usepackage{subcaption} 
\usepackage{makecell}
\usepackage{tabularx}
\usepackage{array}
\usepackage{algorithm, algorithmic}

\definecolor{linkblue}{rgb}{0.10, 0.40, 0.70}
\hypersetup{
colorlinks=true,
citecolor=linkblue,
linkcolor=linkblue,
urlcolor=linkblue,
}

\usepackage{wrapfig}
\usepackage{graphicx}
\usepackage{caption}
\usepackage{multirow}

\newcommand{\modelname}{{\textsc{CATok}}}

\definecolor{darkgreen}{rgb}{0.0, 0.5, 0.0}

\definecolor{chenyu}{rgb}{1.000, 0.322, 0.576}

\title{Rethinking Causal Action Tokenization with Conditional Annealing in Flow Matching}

\author{%
      Chenyu Zhang$^{1,2}$\thanks{\noindent Equal contribution. \textsuperscript{\dag}Corresponding author.}\quad
      Yuhang Cao$^{1}$\footnotemark[1]\quad
      Daru Du$^{1,2}$\quad
      Yingxi Lu$^{1,2}$\quad
      Jing Shao$^{1}$
      \\
      \textbf{
      Ruoqu Chen$^{1,2}$\quad
      Jiajun Liu$^{1,2}$\quad
      Liu Cao$^{1,2}$\quad
      Yicheng Liu$^{1}$\quad
      Hang Zhao$^{1,2}$\quad
      Mengdi Xu$^{1,2\dagger}$
      }
      \\[1ex]
      $^{1}$IIIS, Tsinghua University \hspace{1em}$^{2}$ Shanghai Qizhi Institute\\
      \texttt{\{cyzhang21@mails, caoyh24@mails, xumd@mail\}.tsinghua.edu.cn}\\
      \url{https://chenyuzhangx.github.io/CATok/}
  }

\begin{document}

\maketitle

\begin{abstract}
Autoregressive Vision-Language-Action (VLA) models offer a scalable path to robot learning, yet existing action tokenizers treat tokenization as a compression problem, producing representations that are semantically misaligned with the autoregressive backbone.
We propose \textbf{\modelname}, a causal action tokenizer that reframes tokenization as a causally structured generative process.
\modelname~introduces a \emph{conditional annealing mechanism} that extracts action tokens by progressively annealing a flow-matching process: each token is conditioned on all preceding tokens and encodes the residual reconstruction signal at a specific noise level, establishing a coarse-to-fine causal token space whose generative semantics are structurally aligned with autoregressive modeling.
A \emph{token-conditioned flow-matching decoder} built on Multimodal Diffusion Transformer (MMDiT) reconstructs continuous action chunks from these discrete tokens with the precision of hybrid diffusion-head architectures.
This discrete bottleneck enforces \emph{knowledge insulation} by design, cleanly separating high-level semantic reasoning from low-level motor execution without requiring explicit attention masking.
Extensive evaluations across three simulation benchmarks and real-world robotic manipulation tasks demonstrate that \modelname~consistently surpasses existing tokenization methods in both reconstruction fidelity--compression tradeoff and inference efficiency, while improving VLA task success rate and training efficiency, establishing a high-performance, scalable foundation for purely autoregressive VLA systems.
\end{abstract}
\section{Introduction}

Robot foundation models have shown strong generalization across diverse manipulation tasks~\cite{brohan2023rt2, kim2024openvla, driess2023palm, ghosh2024octo,  black2024pi_0, intelligence2025pi_05, kim2026cosmos, bi2025motus}.
Vision-Language-Action (VLA) architectures have emerged as a particularly promising paradigm, jointly modeling language instructions, visual observations, and motor commands within a unified policy~\cite{brohan2023rt2,kim2024openvla, black2024pi_0, intelligence2025pi_05}.
Existing VLAs broadly fall into two families: \emph{purely autoregressive models} that predict discretized action tokens using the same next-token objective as language models~\cite{brohan2023rt2,kim2024openvla,qu2025spatialvla}, and \emph{hybrid models} that pair an autoregressive language backbone with a continuous diffusion or flow-matching action head~\cite{black2024pi_0,intelligence2025pi_05}. Hybrid models achieve precise continuous control and avoid the bottleneck of naive action discretization, but they decouple action generation from the autoregressive backbone, limiting the unified token-level modeling and scalable LLM training infrastructure that make purely autoregressive VLAs appealing~\cite{cen2025worldvlaautoregressiveactionworld, liu2025unified, hu2026arvlatrueautoregressiveaction}.

A central challenge for purely autoregressive VLAs is therefore how to tokenize continuous robot actions.
The simplest strategy is per-dimension uniform binning~\cite{brohan2023rt2,kim2024openvla,qu2025spatialvla}, where each scalar action dimension is independently mapped to a discrete bin. 
This produces long token sequences of length $O(D{\times}H)$, ignores correlations across action dimensions, and provides no meaningful causal structure within an action chunk. Recent action tokenizers instead compress entire action chunks into short discrete sequences~\cite{pertsch2025fast,liu2025faster,liu2026oat}, but they remain imperfectly matched to autoregressive generation. 
FAST~\cite{pertsch2025fast} uses DCT compression and BPE to obtain compact codes, yet its frequency-ordered coefficients are not naturally aligned with left-to-right autoregressive generation, and its variable-length outputs can make decoding brittle~\cite{liu2026oat}. 
FASTer~\cite{liu2025faster} learns fixed-length RVQ codes with strong reconstruction fidelity, but optimizing a codebook for reconstruction alone does not ensure compatibility with the language-model backbone. 
OAT~\cite{liu2026oat} introduces a left-to-right ordering through nested dropout, but the resulting token positions lack semantic grounding, since each position has no principled correspondence to information granularity or generative stage.

These limitations point to a deeper issue: \emph{existing action tokenizers mainly encode \textbf{what} action should be reconstructed, but not \textbf{how} that action should be generated.} In generative modeling, denoising processes naturally expose a hierarchy of abstraction: high-noise steps capture coarse global structure, while low-noise steps refine fine-grained details~\cite{yue2024diti}. If action tokens were aligned with this hierarchy, then a token sequence could provide a causally ordered, coarse-to-fine representation of continuous control. Such a representation would be better matched to autoregressive generation: early tokens would specify global action structure, and later tokens would refine local motor details. However, existing action tokenizers do not explicitly exploit this correspondence between denoising hierarchy and left-to-right token generation.

\begin{wrapfigure}{r}{0.5\textwidth}
    \centering
    \includegraphics[width=\linewidth]{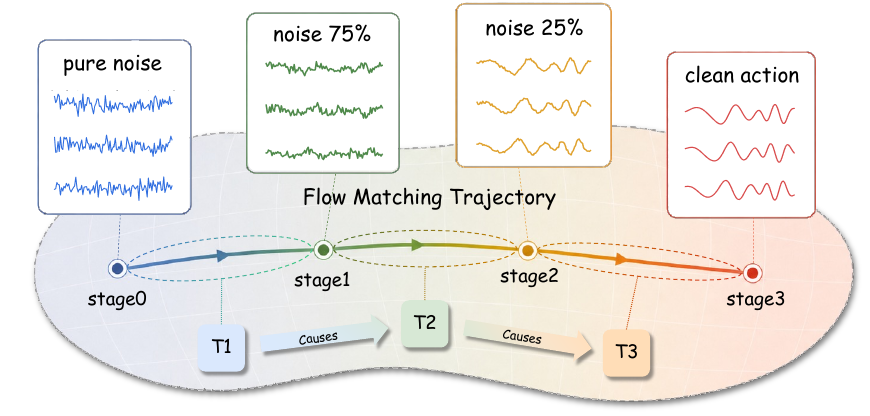}
    \caption{
    \textbf{Flow-Matching to Action Tokens.}
    \modelname~grounds discrete action tokens in the continuous flow matching trajectory. Each token is learned as a stage-wise information increment, imbuing the sequence with a natural coarse-to-fine hierarchy. By construction, these increments follow the temporal evolution of the flow, ensuring a causally-ordered structure that seamlessly aligns with autoregressive generation.}
    \label{fig:teaser}
\end{wrapfigure}
Inspired by these observations, we introduce \modelname, a causal action tokenizer that aligns action tokenization with the generative structure of autoregressive models. 
\modelname{} represents action chunks as compact discrete tokens and reconstructs them using an MMDiT-based flow-matching decoder~\cite{esser2024scaling,liu2022rectifiedflow}, achieving the precision of diffusion-based action heads~\cite{black2024pi_0} while remaining fully compatible with autoregressive VLA training. To structure the token space, we employ a conditional annealing process in which each token is generated conditioned on its predecessors and corresponds to a stage of the denoising process, resulting in a causally ordered coarse-to-fine representation of actions~\cite{yue2024diti} (illustrated in Figure~\ref{fig:teaser}). Experiments on three simulation benchmarks show that \modelname{} achieves the highest success rates across all benchmarks, while delivering a $1.7\times$ faster VLA inference speed and a $3.6\times$ stronger reconstruction fidelity--compression balance than FAST.

Our contributions are three-fold:
\begin{itemize}
  \item \textbf{Causal action tokenization via conditional annealing.}
  We formulate action tokenization as a sequential generative process via conditional annealing, producing tokens that follow a coarse-to-fine, causally ordered structure aligned with autoregressive modeling.

  \item \textbf{Token-conditioned flow matching decoder.}     
  We introduce an MMDiT-based~\cite{esser2024scaling} flow-matching decoder~\cite{liu2022rectifiedflow} that reconstructs continuous action chunks from compact discrete tokens, combining the control fidelity of continuous generative action heads with the training compatibility of purely autoregressive VLAs.

  \item \textbf{Strong empirical performance.}
  \modelname{} consistently outperforms prior methods such as FAST~\cite{pertsch2025fast} and OAT~\cite{liu2026oat} across multiple simulation and real-world benchmarks in reconstruction fidelity--compression tradeoff, inference efficiency and VLA task success rate.
\end{itemize}
\section{Related Works}
\textbf{Vision-Language-Action Models.}
Benefiting from the strong capability of pretrained VLMs~\cite{chen2023pali, beyer2024paligemma, qwen3technicalreport} to understand images and language instructions, VLAs demonstrate remarkable performance and generalization in robotic manipulation tasks~\cite{driess2023palm, zitkovich2023rt}. Existing VLA frameworks primarily adopt two paradigms for action integration: discrete tokenization for autoregressive generation~\cite{kim2024openvla, qu2025spatialvla, pertsch2025fast, liu2025unified, goyal2025vla}, and continuous regression via an auxiliary action head conditioned on the VLM's latent representations~\cite{black2024pi_0, kim2025fine, intelligence2025pi_05, shukor2025smolvla}.
Despite the fact that discrete-token autoregressive paradigms incur excessive inference delays, recent works reveal a contrasting boost in training performance. FAST~\cite{pertsch2025fast} demonstrates that employing a tokenizer that effectively compresses action chunks can significantly enhance training efficiency. Moreover, fine-tuning VLMs via discrete tokens, rather than continuous action heads, has been shown to better preserve the rich semantics of the model~\cite{driess2025knowledge}. Together, these findings underscore the critical importance of designing an optimal discrete action tokenizer.

\textbf{Action Tokenization.}
Inherently, an action chunk is represented as a two-dimensional matrix spanning both temporal and spatial (action) dimensions. To process such structures, researchers have proposed a variety of action tokenization strategies, ranging from rule-based methods to data-driven approaches.
Early binning-based action discretization~\cite{kim2024openvla, qu2025spatialvla, zitkovich2023rt} introduces substantial token redundancy, severely compromising both training and inference speeds. More critically, this naive mapping ignores inter-token dependencies, limiting its effectiveness under the standard next-token-prediction framework.
Compression-based methods (e.g., DCT~\cite{pertsch2025fast} and B-splines~\cite{zhou2025beast}) were subsequently introduced to compress each action dimension independently along the temporal axis. While these approaches substantially improve the compression ratio---particularly at high control frequencies---and boost training efficiency, they still fail to capture correlations across action dimensions.
Most recently, learning-based tokenizers~\cite{liu2025unified, dong2026actioncodec, huang2026mimic} have emerged to model the spatiotemporal dependencies of actions, with a growing focus on the hierarchical organization and underlying structure of the token space. For instance, RVQ-based approaches~\cite{liu2025faster, wang2025vq} capture multi-level hierarchies through residual quantization, whereas OAT~\cite{liu2026oat} enforces structural integrity in the token space by applying random token masking in the action decoder.\par
While promising, these tokenization paradigms leave room for further refinement in balancing high-quality reconstruction with token-level causal structures. A key objective, therefore, is to develop a training formulation that naturally yields a structured latent topology.
To this end, we propose a method that transfers the temporal causal structure inherent in Flow Matching~\cite{yue2024diti, wang2025selftok, wangselftok} into the token space, thereby inducing causality among tokens. This design is empirically validated to be effective for autoregressive modeling.
\section{Method}
                             
In this section, we first formalize the action tokenization problem in Section~\ref{sec:problem_formulation}; then, we describe the \modelname~architecture in Section~\ref{sec:catok_architecture} and detail its training and inference procedures in Section~\ref{sec:catok_train_infer}. Finally, Section~\ref{sec:vla-catok} presents VLA-\modelname, which integrates the proposed tokenizer into an autoregressive VLA backbone.

\subsection{Problem Formulation} 
\label{sec:problem_formulation}

We study action tokenization for purely autoregressive VLA models. 
Given a continuous action chunk $\mathcal{A} = (a_t,\dots,a_{t+H-1}) \in \mathbb{R}^{H \times d_a}$, the goal is to learn an encoder $\mathcal{E}$ and decoder $\mathcal{D}$ such that
\[
\mathcal{E}(\mathcal{A}) = C = (c_1,\dots,c_K), \quad c_k \in \{1,\dots,|\mathcal{C}|\}, \qquad
\mathcal{D}(C) \approx \mathcal{A},
\]
where $\mathcal{C}$ is a finite codebook and $K$ is the token sequence length. 
The tokenizer is trained to minimize reconstruction error:
\[
\min_{\mathcal{E},\mathcal{D}} \mathbb{E}_{\mathcal{A}}\big[\|\mathcal{D}(\mathcal{E}(\mathcal{A})) - \mathcal{A}\|^2\big].
\]

Once trained and frozen, $\mathcal{E}$ and $\mathcal{D}$ convert action chunks into discrete tokens $C$, which a purely autoregressive VLA policy $\pi_\theta$ predicts conditioned on observations $o_t = (I_t, s_t, l)$. 
The VLA is trained with cross-entropy over tokens:
\[
\min_\theta \mathbb{E}_{(o_t,\mathcal{A})} \Big[-\sum_{k=1}^K \log \pi_\theta(c_k \mid o_t, c_{<k}) \Big].
\]
At inference, $\pi_\theta$ generates $\hat{C} = (\hat{c}_1,\dots,\hat{c}_K)$ autoregressively, and the frozen decoder recovers the continuous action $\hat{\mathcal{A}} = \mathcal{D}(\hat{C})$ for execution.

\subsection{\modelname~Architecture}
\label{sec:catok_architecture}

\begin{figure}
  \centering
  \includegraphics[width=\linewidth]{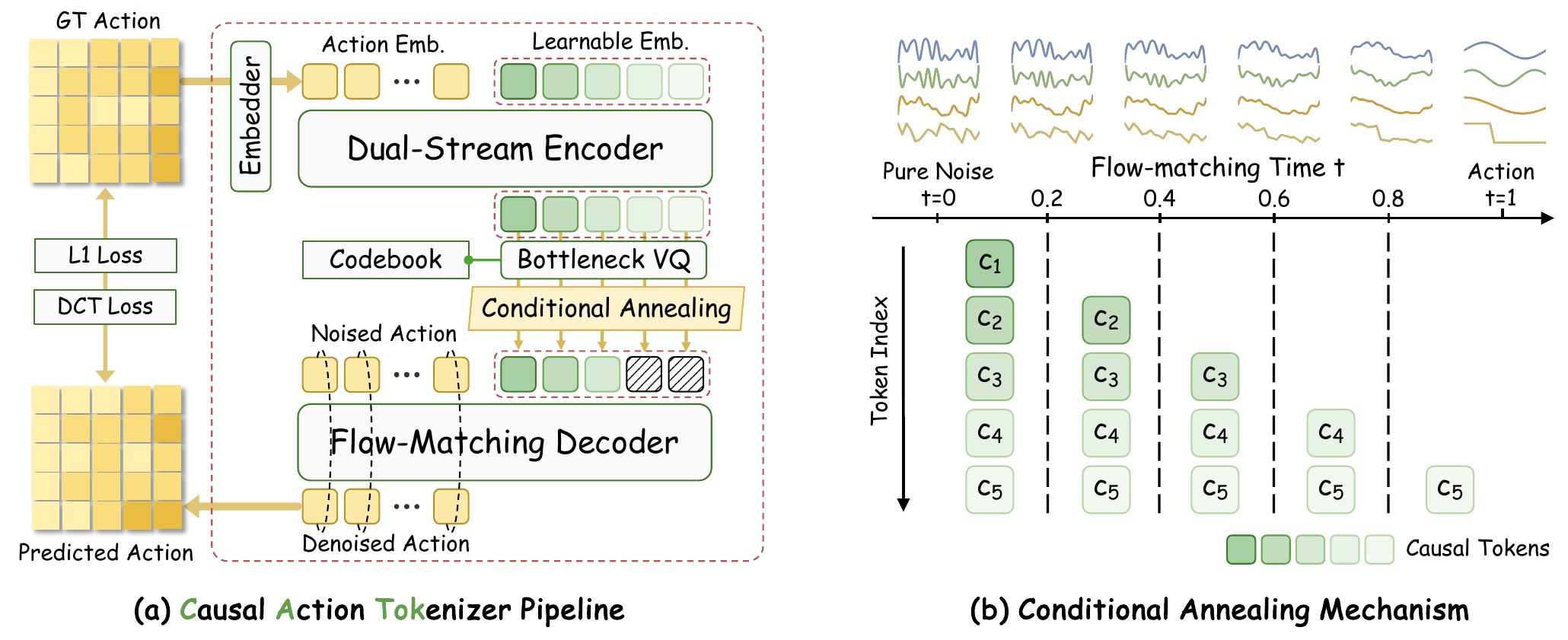}
  \caption{
  \textbf{An Overview of \modelname.}
    (a) \textbf{\modelname~Pipeline.} \modelname~encodes an action chunk with a \emph{Dual-Stream Encoder} into latent queries, discretizes them via a \emph{Bottleneck VQ}, and reconstructs actions using a \emph{Flow-Matching Decoder with Conditional Annealing}.
    (b) \textbf{Conditional Annealing Mechanism} progressively masks the first $\kappa(t)$ token embeddings according to flow matching timestep t: darker colors indicate tokens earlier in the sequence, and lighter colors indicate later tokens, assigning each token to a different stage of the flow trajectory.
}

  \label{fig:catok}
\end{figure}

Our causal action tokenizer, \modelname, encodes action chunks using a \textbf{Dual-Stream Action Encoder} that maps them into a compact set of latent query tokens via co-attention. 
These latent tokens are then discretized through a \textbf{Bottleneck Vector Quantizer} into a finite codebook. 
Finally, a \textbf{Flow-Matching Decoder with Conditional Annealing} reconstructs high-fidelity action trajectories while enforcing a coarse-to-fine causal structure over the tokens.

\textbf{Dual-Stream Action Encoder.} 
We first normalize the raw action chunk using quantile-based normalization to mitigate the impact of outliers and improve numerical stability. 
Instead of flattening or 1D patchification, we encode $\mathcal{A}_{t:t+H}$ using a 2D convolutional neural network (CNN) with weight normalization~\cite{salimans2016weight}, which captures local correlations along both temporal and action dimensions. 
We further add learnable 2D positional embeddings to preserve the spatiotemporal structure, yielding structured embeddings $\mathbf{E}_{\text{act}}$.

In parallel, we initialize $K$ learnable query embeddings $\mathbf{Q}^{(0)} \in \mathbb{R}^{K \times D}$ with independent positional embeddings. 
We then feed both $\mathbf{E}_{\text{act}}$ and $\mathbf{Q}^{(0)}$ into a symmetric multi-modal transformer inspired by MMDiT~\cite{esser2024scaling}, where co-attention enables bidirectional interaction between the two token sets (Figure~\ref{fig:catok}(a)). 
In this process, the action embeddings provide structured context, while the query tokens aggregate information into a compact latent space.
After $L$ layers, we retain the updated query stream $\mathbf{Q} = (\mathbf{q}_1, \dots, \mathbf{q}_K) \in \mathbb{R}^{K \times D}$ as the final representation of the action chunk.

\textbf{Bottleneck Vector Quantizer.} 
To obtain a compact discrete representation, we adopt the \emph{factorized codes} design of ViT-VQGAN~\cite{yu2021vector}. Each latent vector $\mathbf{q}_k \in \mathbb{R}^D$ is first projected into a lower-dimensional space through a linear bottleneck $\psi:\mathbb{R}^D \rightarrow \mathbb{R}^{D'}$ with $D' < D$. The projected embedding $\psi(\mathbf{q}_k)$ is then quantized against a learnable codebook $\mathcal{C} = \{\mathbf{e}_1, \dots, \mathbf{e}_{|\mathcal{C}|}\} \subset \mathbb{R}^{D'}$ by selecting its nearest neighbor indexed $c_k$ in Euclidean distance:
\begin{equation}
  c_k = \mathop{\arg\min}_{j \in \{1,\dots,|\mathcal{C}|\}}\; \bigl\|\mathbf{e}_j - \psi(\mathbf{q}_k)\bigr\|_2,
\end{equation}
Finally, a post-projection layer $\phi: \mathbb{R}^{D'} \rightarrow \mathbb{R}^D$ is applied to reconstruct the quantized embedding in the original latent space, $\hat{\mathbf{q}}_k = \phi({\mathbf{e}}_{c_k})$. This factorized design substantially improves codebook utilization by decoupling the codebook dimension from the model's hidden dimension.

\textbf{Flow-Matching Decoder with Conditional Annealing.}
Prior work suggests that diffusion/flow timesteps naturally correspond to different levels of abstraction, from coarse global structure to fine-grained details~\cite{yue2024diti}. 
We exploit this property to ground discrete tokens in specific generative stages via a \emph{conditional annealing} mechanism (as illustrated in Figure~\ref{fig:catok} (b)).

Let $x_0\sim\mathcal{N}(0,I)$ and $x_1=\mathcal{A}$ denote Gaussian noise and the target action chunk respectively. 
Following rectified flow, we define the interpolation path
$
x_t=(1-t)x_0+t x_1
$
with target velocity $x_1-x_0$. 
Instead of conditioning the decoder on all token embeddings throughout the entire trajectory, we progressively mask the first $\kappa(t)$ tokens according to
\begin{equation}
    \kappa(t)=\lfloor tK\rfloor,\quad t\in[0,1],
    \label{eq:annealing_mask}
\end{equation}
so that only the active subset $\hat{\mathcal{Q}}_{>\kappa(t)}$ is provided as conditioning at time $t$. 
The MMDiT-based velocity network~\cite{esser2024scaling} is trained with the flow-matching objective
\begin{equation}
    \mathcal{L}_{\mathrm{FM}}
    =
    \mathbb{E}_{t,x_0,x_1}
    \left[
    \left\|
    v_\theta(x_t,t,\hat{\mathcal{Q}}_{>\kappa(t)})
    -
    (x_1-x_0)
    \right\|_1
    \right].
\end{equation}

This schedule forces different tokens to be useful at different stages of the generation process. 
At early stages, when $x_t$ is dominated by noise, the active tokens must provide coarse global guidance;
 as $t$ increases and the trajectory approaches the target action, fewer tokens remain active, encouraging later tokens to specialize in residual, fine-grained corrections. 
Consequently, each token embedding acts as a discrete information increment between adjacent flow-matching stages, inducing a causally ordered coarse-to-fine hierarchy that is naturally aligned with autoregressive action generation.

\subsection{\modelname~Training and Inference}
\label{sec:catok_train_infer}

\textbf{Training Scheme.} 
The model is trained end-to-end with a combination of reconstruction and quantization losses:
\begin{equation}
\mathcal{L} = \mathcal{L}_{\text{FM}} + \alpha \mathcal{L}_{\text{smooth}} + \beta \mathcal{L}_{\text{VQ}},
\end{equation}
where $\alpha, \beta$ are hyperparameters that balance the three terms.

Specifically, the reconstruction loss integrates constraints in both the time and frequency domains. It consists of a time-domain $L_{\text{FM}}$ loss which together with a frequency-domain $L_{\text{smooth}}$ loss computed via the discrete cosine transform (DCT):
\begin{equation}
\mathcal{L}_{\text{smooth}} = \mathbb{E}_{t, x_0, x_1} \bigl\| \mathrm{DCT}(x_1-x_0) - \mathrm{DCT}(v_\theta(x_t, t, \hat{\mathcal{Q}}_{>\kappa(t)})) \bigr\|_1,
\end{equation}
The DCT term explicitly penalizes spectral discrepancies, thereby promoting smoother and more temporally consistent predictions over long horizons.
Complementing the reconstruction objective, the quantization loss adopts the standard commitment formulation:
\begin{equation}
\mathcal{L}_{\text{VQ}} = \sum_{i=1}^{K} \bigl\|\psi(\mathbf{q}_i) - \mathrm{sg}(\hat{\mathbf{q}}_i)\bigr\|_2^2,
\end{equation}
where $\mathrm{sg}(\cdot)$ denotes the stop-gradient operator~\cite{van2017neural}.
To further stabilize training and mitigate codebook collapse, codebook entries are updated using exponential moving averages (EMA) of the assigned embeddings, with rarely used entries, identified by low EMA usage, periodically reinitialized from randomly sampled embeddings in the current batch~\cite{van2017neural,lee2022rq}.

\textbf{Detokenization.}
Given action tokens $C=(c_1,\ldots,c_K)$, each $c_i$ is mapped to embedding $\hat{\mathbf{q}}_i \in \mathcal{C}$, yielding $\hat{\mathbf{Q}}=(\hat{\mathbf{q}}_1,\ldots,\hat{\mathbf{q}}_K)$. The action is then decoded via flow matching by solving an ODE:
\begin{equation}
      \mathbf{x}_1 = \mathbf{x}_0 + \int_{0}^{1} v_\theta\!\left(\mathbf{x}_t,t \mid \tilde{\mathbf{Q}}_t\right)dt, \qquad \mathbf{x}_0 \sim \mathcal{N}(\mathbf{0},\mathbf{I}),
\end{equation}
where $v_\theta$ denotes the velocity field and $\tilde{\mathbf{Q}}_t = (\mathbf{0}, \ldots, \mathbf{0}, \hat{\mathbf{q}}_{\kappa(t)+1}, \ldots, \hat{\mathbf{q}}_K)$ retains only the last $K - \kappa(t)$ embeddings via a conditional annealing scheme. 

\subsection{VLA-\modelname} 
\label{sec:vla-catok}
\begin{figure}
  \centering
  \includegraphics[width=\linewidth]{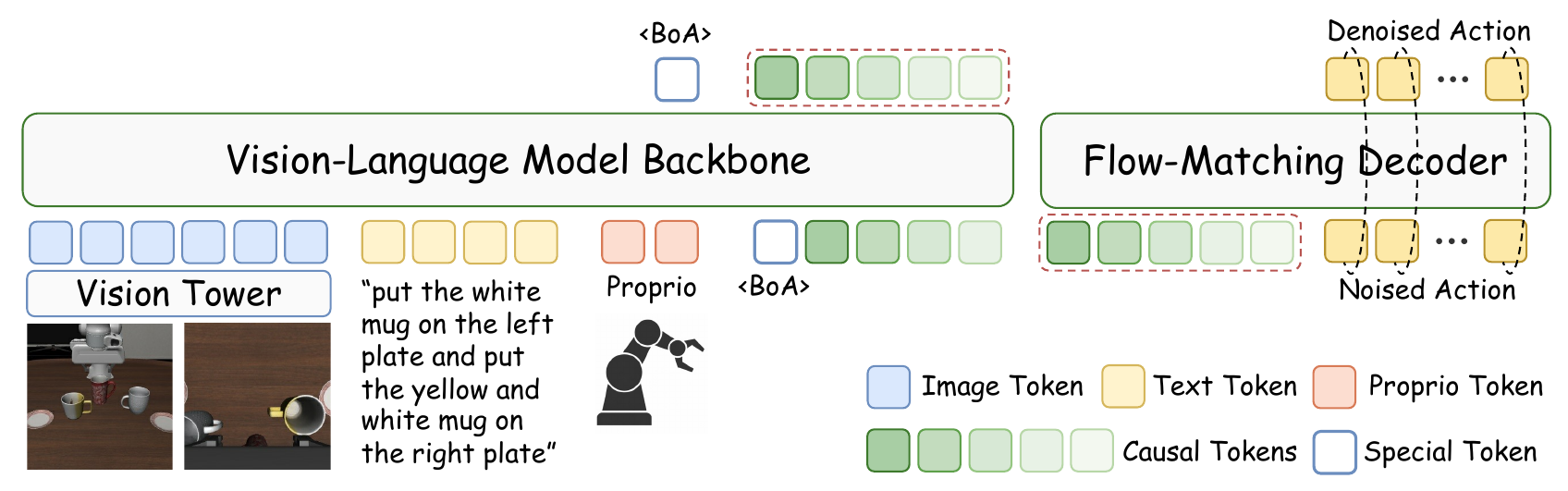}
  \caption{\textbf{An Overview of VLA-\modelname}. VLA-\modelname~feeds visual, language, and proprioceptive tokens into a VLM backbone to autoregressively predict causal action tokens. These tokens are mapped via the learned codebook and processed by a frozen MMDiT flow-matching decoder to produce the final action in a single denoising step. Discrete tokens serve as the sole interface, naturally insulating the VLM's pretrained knowledge from action-specific gradients.}
  \label{fig:vla-catok}
\end{figure}
To evaluate \modelname{} in robotic policy learning, we integrate it into a standard autoregressive VLA architecture~\cite{kim2024openvla, pertsch2025fast}. 
As shown in Figure~\ref{fig:vla-catok}, VLA-\modelname{} takes the current image $I_t$, language instruction $l$, and optional proprioceptive state $s_t$ as input, and autoregressively predicts a sequence of action tokens $C=(c_1,\ldots,c_K)$. 
Each predicted token $c_i$ is mapped to its VQ embedding $\hat{\mathbf{q}}_i$ through the learned codebook, forming $\hat{\mathbf{Q}}=(\hat{\mathbf{q}}_1,\ldots,\hat{\mathbf{q}}_K)$. 
A frozen MMDiT flow-matching decoder then converts $\hat{\mathbf{Q}}$ into the final continuous action chunk in a single decoding step.

During VLA training, the \modelname{} tokenizer and decoder are kept frozen, and only the autoregressive VLA backbone is updated. 
The backbone is trained with teacher-forced next-token prediction using a cross-entropy loss over \modelname{} tokens:
\begin{equation}
    \mathcal{L}_{AR} = -\sum^K_{i=1} {\log {p_{\theta} \left( c_{i} | I_t, s_t, l, c_{<i} \right)}}
\end{equation}
Unlike approaches that supervise actions directly and thus propagate action gradients back into the VLM backbone, our framework uses discrete tokens as the interface between the two modules and freezes the pretrained decoder, inherently insulating knowledge between the two modules. This substantially preserves the pretrained knowledge embedded in the VLM, thereby strengthening its instruction-following capability~\cite{driess2025knowledge}.
\section{Experiments}

To comprehensively evaluate the effectiveness of \modelname{}, we conduct a series of experiments covering both the tokenizer itself and its integration into downstream VLA policies.
Section~\ref{sec:experimental-setup} introduces the tokenizer baselines, VLA backbones, and evaluation benchmarks, including both simulation and real-world settings.
Section~\ref{sec:vla-training} examines whether \modelname{} improves downstream VLA performance and training efficiency. 
Section~\ref{sec:causal-structure} analyzes whether \modelname{} produces causally ordered tokens with generative semantics. 
Finally, Section~\ref{sec:intrinsic-properties} studies the intrinsic properties of the tokenizer, including reconstruction fidelity, compression efficiency, and tokenization latency. 

\subsection{Experimental Setup}
\label{sec:experimental-setup}

\textbf{Tokenizer Baselines.}
We compare \modelname{} with three other representative action tokenizers for autoregressive policy learning: 
\textbf{BIN}~\cite{kim2024openvla}, which uniformly discretizes each action dimension into 256 bins; 
\textbf{FAST}~\cite{pertsch2025fast}, which applies DCT-based action compression followed by BPE tokenization; and 
\textbf{OAT}~\cite{liu2026oat}, a learning-based VQ-VAE action tokenizer that enforces token-level causal structure through prefix-based masking and decoding.
All tokenizers are integrated into the same autoregressive policy framework, and we further analyze their intrinsic token properties in Section~\ref{sec:intrinsic-properties}.

\textbf{Simulation Benchmarks.}
We evaluate downstream policy performance on three simulation benchmarks: \textbf{LIBERO}~\cite{liu2023liberobenchmarkingknowledgetransfer}, \textbf{SimplerEnv}~\cite{li2024evaluatingrealworldrobotmanipulation}, and \textbf{RoboTwin 2.0}~\cite{chen2025robotwin20scalabledata}.
LIBERO consists of four official suites covering diverse long-horizon manipulation tasks.
SimplerEnv evaluates zero-shot generalization on real-world-inspired robotic manipulation tasks using the WidowX platform.
RoboTwin 2.0 provides large-scale embodied manipulation benchmarks under both clean and randomized environments, covering 50 tasks with substantial scene diversity.

\begin{figure}[h]
  \centering
  \begin{subfigure}[t]{0.32\textwidth}
    \centering
    \includegraphics[width=\linewidth]{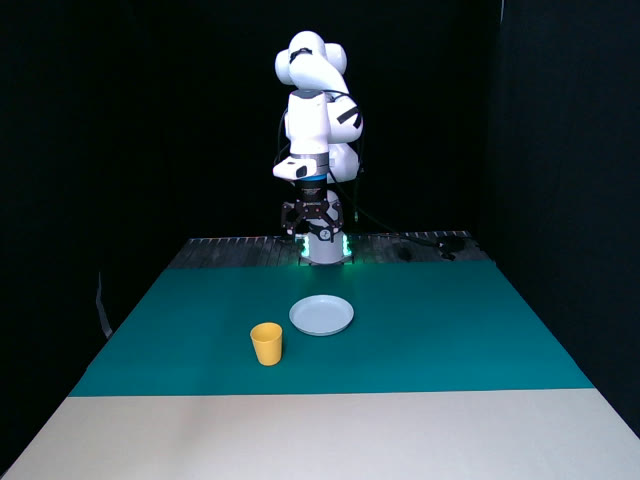}
    \caption{Pick-Spatial}
    \label{fig:catok_pick_spatial}
  \end{subfigure}\hfill
  \begin{subfigure}[t]{0.32\textwidth}
    \centering
    \includegraphics[width=\linewidth]{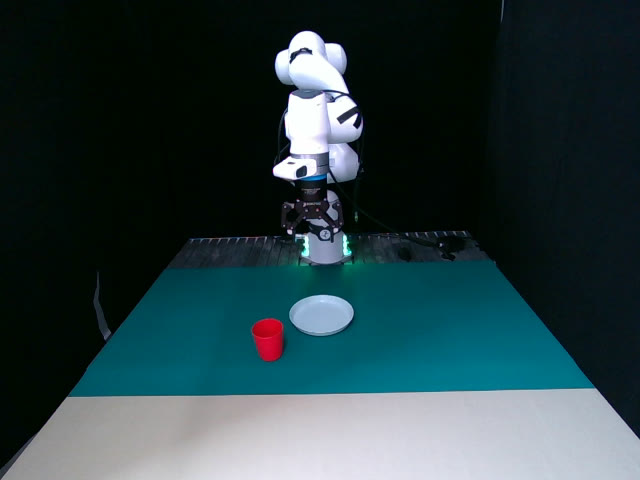}
    \caption{Pick-Color}
    \label{fig:catok_pick_color}
  \end{subfigure}\hfill
  \begin{subfigure}[t]{0.32\textwidth}
    \centering
    \includegraphics[width=\linewidth]{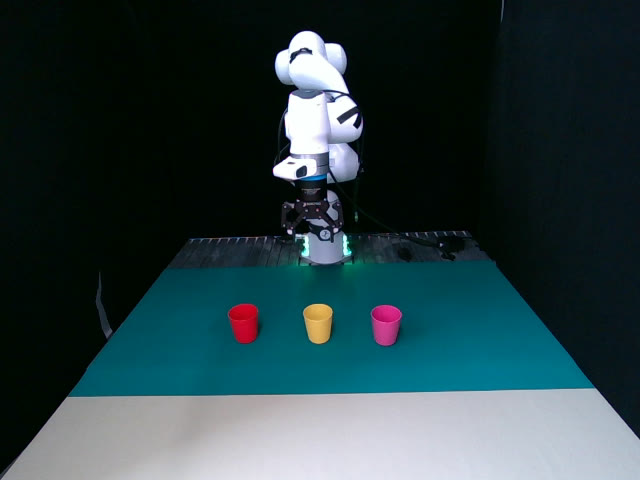}
    \caption{Stack-Long}
    \label{fig:catok_stack_long}
  \end{subfigure}
  \caption{
    \textbf{Real-world scene layouts for the three task suites.}
    From left to right: \textit{Pick-Spatial}, testing spatial generalization by varying cup and plate positions; 
    \textit{Pick-Color}, testing color generalization with different unseen cup colors; 
    and \textit{Stack-Long}, testing long-horizon manipulation under different spatial layouts.
}
  \label{fig:catok_scene_layouts}
\end{figure}

\textbf{Real-World Benchmark.}
We further evaluate \modelname{} on a real Franka manipulator under three task suites, as shown in Fig.\ref{fig:catok_scene_layouts}: \textit{Pick-Spatial}, \textit{Pick-Color}, and \textit{Stack-Long}. 
The tasks are designed to test spatial generalization, color generalization, and long-horizon task completion, respectively.

For training, we collect 362 trajectories for \textit{Pick-Cups} task and 220 trajectories for \textit{Stack-Cups} task.
Using Qwen3.5-VL as the shared backbone, we train QwenCATok and QwenFAST with the same training protocol and different action tokenizers.
To ensure a comparable reconstruction quality between tokenizers, CATok is pretrained on the collected real-robot data together with InternA1 joint data, yielding reconstruction accuracy comparable to FAST.

\textbf{Training and Evaluation Protocols.}
For LIBERO, we fine-tune \textbf{$\pi_0$-FAST-base}~\cite{pertsch2025fast} on all four suites and report the average success rate over 40 tasks with 50 rollouts per task.
For the more challenging SimplerEnv and RoboTwin 2.0 benchmarks, all methods are trained from scratch using \textbf{Qwen3-VL-4B-Instruct}~\cite{qwen3technicalreport} as the shared VLM backbone to ensure fair comparison.
In SimplerEnv, policies are trained on Bridge~\cite{walke2023bridgedata} and evaluated zero-shot on the four WidowX tasks following the official protocol.
In RoboTwin 2.0, we train on the official clean and randomized datasets, and evaluate on all 50 tasks under both clean and randomized settings with 20 rollouts per task.
Additional implementation and evaluation details are provided in the Appendix~\ref{appendix:implementation}.

\begin{table}[t]
\centering
\caption{\textbf{Comparison on robotic manipulation  simulation benchmarks.} \modelname~achieves the best overall performance across all benchmarks, with particularly significant improvements on long-horizon tasks such as Long in LIBERO and \textit{StackGreenCubeOnYellowCube} in SimplerEnv, demonstrating the effectiveness of causal action tokenization for modeling long-range temporal dependencies.}
\label{tab:success-rate}
\setlength{\tabcolsep}{5pt}
\renewcommand{\arraystretch}{1.25}
\resizebox{\linewidth}{!}{%
\begin{tabular}{
  l
  c c c c c
  c c c c c
  c c
}
\toprule
& \multicolumn{5}{c}{\text{LIBERO}}
& \multicolumn{5}{c}{\text{SimplerEnv}}
& \multicolumn{2}{c}{\text{RoboTwin 2.0}} \\
\cmidrule(lr){2-6}\cmidrule(lr){7-11}\cmidrule(lr){12-13}
\text{Method}
  & Spatial & Object & Goal & Long & Avg.
  & Spoon  & Carrot & Stack & Eggplant & Avg.
  & Clean  & Randomized \\
\midrule
BIN
  & 0.586 & 0.878 & 0.680 & 0.604 & 0.687
  & \textbf{0.542} & 0.333 & 0.208 & \textbf{0.708} & 0.448
  & 0.213   & 0.221 \\
FAST
  & 0.960 & \textbf{0.998} & \textbf{0.962} & 0.901 & 0.955
  & 0.500 & 0.375 & 0.375 & 0.625 & 0.469
  & 0.478 & 0.478 \\
OAT
  & 0.428 & 0.876 & 0.704 & 0.276 & 0.571
  & 0.417 & 0.208 & 0.167 & 0.667 & 0.365
  & 0.229   & 0.233 \\
\modelname
  & \textbf{0.978} & 0.994 & 0.954 & \textbf{0.910} & \textbf{0.959}
  & 0.458 & \textbf{0.417} & \textbf{0.542} & 0.542 & \textbf{0.490}
  & \textbf{0.489}   & \textbf{0.531} \\
\bottomrule
\end{tabular}}
\end{table}

\subsection{Can \modelname~ Help VLA Training?}
\label{sec:vla-training}

\textbf{\modelname~consistently improves VLA success rates across simulation benchmarks.}
As shown in Table~\ref{tab:success-rate}, \modelname~consistently outperforms all baseline tokenizers across the three simulation benchmarks. 
Notably, it exhibits particularly pronounced gains on long-horizon manipulation tasks. In particular, on the \textit{StackGreenCubeOnYellowCube} task in SimplerEnv, \modelname~improves the success rate by more than 16.7\% relative to the strongest baseline, highlighting its superior ability to model extended sequences of coordinated actions. These results suggest that introducing causal ordering into action tokenization enables autoregressive policies to capture long-range temporal dependencies more effectively, which is crucial for complex long-horizon manipulation tasks.

\begin{wrapfigure}[19]{r}{0.6\textwidth}
  \centering
  \vspace{-7pt}
  \hspace{-8pt}
  \includegraphics[width=\linewidth]{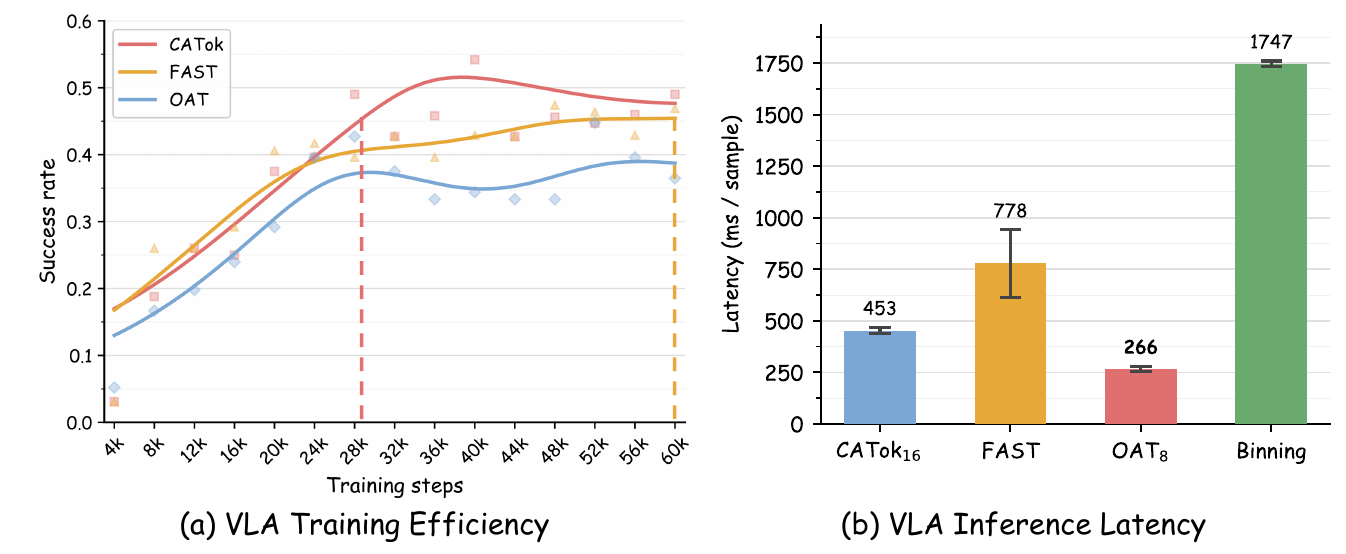}
  \vspace{-6pt}
  \caption{(a) Training efficiency on the SimplerEnv benchmark. Curves are Gaussian-smoothed to reduce evaluation stochasticity without changing the relative ranking of different methods. Notably, \modelname{} reaches the best baseline performance using only around 50\% of the training steps. (b) VLA inference latency. By jointly considering autoregressive generation and decoding overhead, \modelname{} achieves near-minimal end-to-end inference latency while attaining the best success rate among all methods.}
  \label{fig:training-efficiency}
\end{wrapfigure}

\textbf{\modelname~improves both VLA training efficiency and inference speed.}
Figure~\ref{fig:training-efficiency} (a) compares the success rates of \modelname~and baseline tokenizers across training steps, highlighting the superior training efficiency of \modelname. \modelname~consistently achieves higher success
rates, particularly during the later stages of training, indicating both faster convergence and more stable learning dynamics. Notably, \modelname~matches the best performance achieved by FAST with only approximately 50\% steps, demonstrating substantially improved training efficiency. 
We attribute this improvement primarily to the causal token structure analyzed in Section~\ref{sec:causal-structure}.
By organizing action chunks into causally ordered tokens, \modelname~naturally aligns the tokenizer with the autoregressive policy objective, enabling more effective stage-wise action modeling rather than learning from weakly structured discrete codes.
Consequently, during autoregressive generation, each predicted token provides more informative guidance for subsequent token prediction and trajectory generation, consistently improving both training efficiency and policy performance across diverse tasks.
As shown in Figure~\ref{fig:training-efficiency} (b), \modelname{} also exhibits highly competitive inference efficiency. By jointly considering autoregressive token generation and tokenizer decoding overhead, it achieves near-minimal end-to-end VLA inference latency among all compared methods.

\begin{wraptable}{l}{0.6\textwidth}
  \centering
  \vspace{-8pt}
  \setlength{\tabcolsep}{7pt}
  \renewcommand{\arraystretch}{1.15}
  \begin{tabular}{lccc}
    \toprule
    Method & Pick-Spatial & Pick-Color & Stack-Long \\
    \midrule
    FAST & 0.450 & 0.300 & 0.275 \\
    \modelname & \textbf{0.650} & \textbf{0.500} & \textbf{0.475} \\
    \bottomrule
  \end{tabular}
  \caption{\textbf{Real-world manipulation results.}
  \modelname~consistently outperforms FAST across spatial, color, and long-horizon generalization tasks.}
  \label{tab:real-world}
  \vspace{-8pt}
\end{wraptable}

\textbf{\modelname~transfers effectively to real-world robots.}
To further evaluate whether the benefits of causal action tokenization extend beyond simulation, we deploy \modelname~on a real-world Franka robot.
As shown in Table~\ref{tab:real-world}, \modelname~consistently outperforms FAST across three task suites covering spatial generalization, color generalization, and long-horizon manipulation.
Notably, \modelname~achieves a consistent 20 percentage-point improvement over FAST across all three settings, demonstrating that the benefits of causal action tokenization transfer to real-world robotic manipulation.

\subsection{Does \modelname~Produce Causal Tokens with Generative Semantics?}
\label{sec:causal-structure}

\textbf{Causal Predictive Ordering of Action Tokens.}
We test whether \modelname{} tokens exhibit a predictive order by training a lightweight action-only autoregressive transformer for each tokenizer and measuring next-token entropy on held-out LIBERO-Bridge Mixture chunks.
At each position $k$, we compute $H_k=-\sum_v p_{k,v}\log p_{k,v}$ from the prediction logits; lower entropy indicates that preceding tokens provide stronger predictive context.

\begin{figure}[t]
  \centering
  \includegraphics[width=\linewidth]{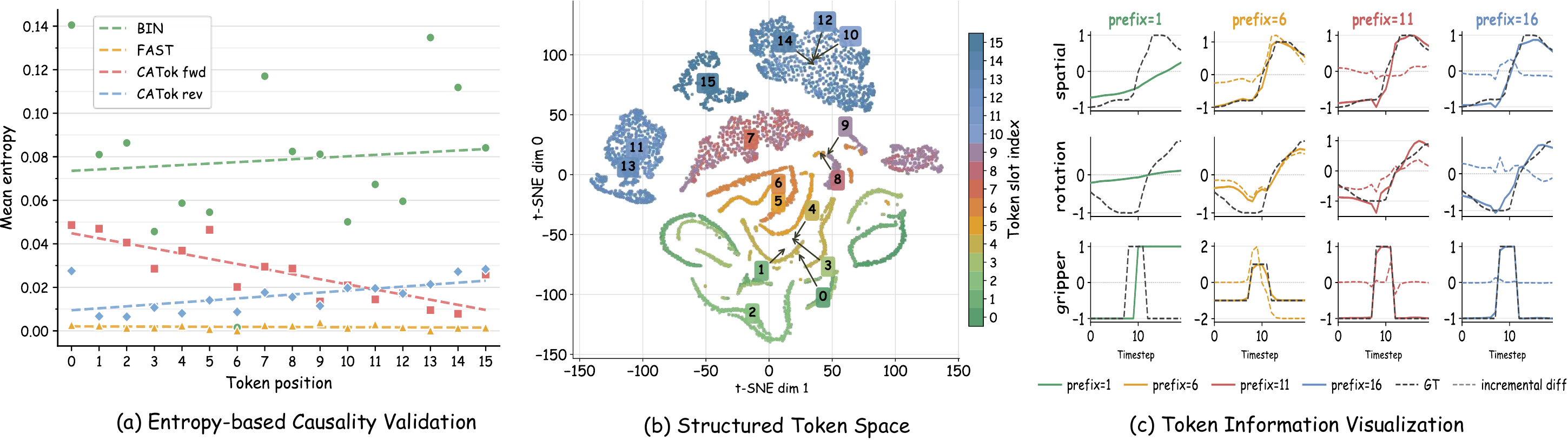}
  \caption{
  \textbf{\modelname~Produces Causal Tokens with Generative Semantics}. 
  \textbf{(a) Causal Ordering} \modelname~'s entropy decreases in the forward order (\emph{CATOK fwd}) and increases in the reverse order (\emph{CATOK rev}), while FAST and Bin show weaker positional structure.
  \textbf{(b) Token embedding geometry.} T-SNE map shows slot-dependent organization, progressing from smooth early-token regions to compact late-token clusters.
  \textbf{(c) Prefix reconstruction.} Increasing token prefixes produce structured trajectory updates, with dashed curves showing incremental contributions.
  }
  \label{fig:token_vis}
\end{figure}

As shown in Figure~\ref{fig:token_vis}(a), \modelname{} exhibits a clear decreasing entropy trend when tokens are predicted in their original order, indicating that later tokens become progressively easier to predict as more prefix tokens are observed.
When predicted in reverse order, the trend is inverted, demonstrating that the learned token ordering is directional rather than arbitrary.
In contrast, FAST remains nearly flat and Bin shows a slight increase, suggesting substantially weaker positional dependency.
These results indicate that \modelname{} learns causally ordered action tokens that are naturally aligned with autoregressive generation.

\textbf{Stage-wise Generative Semantics.}
We next examine whether \modelname~tokens carry generative semantics, i.e., whether different tokens induce distinct and structured changes in the decoded action chunk. 
We use two complementary probes: token-space geometry and prefix reconstruction.\\
\textbf{Token-space geometry.}
We visualize VQ embeddings from 32,768 normalized LIBERO action chunks using t-SNE, with embeddings extracted from a frozen \modelname{} checkpoint ($K{=}16$, codebook size $4096$, $d_{\mathrm{vq}}{=}16$).
As shown in Figure~\ref{fig:token_vis}(b), the embeddings are clearly organized by token slot: early slots form smooth elongated regions, middle slots become more diffuse, and late slots form compact, separated clusters. 
This slot-dependent geometry suggests that \modelname~ tokens are not exchangeable codebook indices, but occupy distinct generative roles that progress from shared motion factors to specialized residual corrections.\\
\textbf{Prefix reconstruction.}
We reconstruct action chunks from token prefixes with $n\in\{1,6,11,16\}$.
As shown in Figure~\ref{fig:token_vis}(c), adding tokens induces coherent trajectory updates, from structural changes in spatial, rotation, and gripper dimensions to finer residual refinements. 
This indicates that \modelname~ tokens serve as meaningful generative increments rather than passive compression codes.

Together, the geometry and reconstruction probes show that \modelname{} learns a coarse-to-fine hierarchy of generative action tokens. 
Earlier tokens capture shared global structure, while later tokens contribute increasingly specialized residual refinements. 
This stage-wise organization complements the causal ordering above and provides a natural inductive bias for autoregressive action generation.

We further validate the causal semantics of individual token positions through
native-decoder intervention experiments, including token swapping and
single-token removal.
These experiments provide direct evidence that different token positions
control distinct stages of action generation.
Detailed experimental settings and results are provided in
Appendix~\ref{appendix:causality-validation}.

\subsection{Intrinsic Properties of \modelname~}
\label{sec:intrinsic-properties}
\textbf{\modelname~balances reconstruction fidelity and compression efficiency.} 
Action tokenization inherently requires balancing reconstruction fidelity and compression efficiency. We evaluate reconstruction quality using Valid Reconstruction Rate (VRR)~\cite{liu2025faster} and measure compactness using Compression Rate (CR). Detailed metric definitions are provided in Appendix~\ref{appendix:metrics}.

As shown in Figure~\ref{fig:tokenizer_properties} (a-c), \modelname{} achieves near-perfect reconstruction fidelity, with VRR comparable to Binning and FAST and substantially higher than OAT under the same token budget. Meanwhile, it provides significantly better compression than Binning and FAST while remaining comparable to OAT, resulting in the best overall fidelity--compression tradeoff.

\textbf{Encoding and decoding latency.}
To evaluate the efficiency of action tokenization during both training and inference, we measure the average encoding and decoding latency during action tokenization. Detailed measurement protocols are provided in Appendix~\ref{appendix:metrics}.
Since \modelname~ substantially outperforms Binning and OAT in downstream VLA performance and achieves performance comparable to FAST, while consistently obtaining slightly higher success rates overall, FAST serves as the most relevant latency baseline. As shown in Figure~\ref{fig:tokenizer_properties} (d-e), compared with FAST, \modelname~ achieves substantially lower encoding and decoding latency while also delivering slightly better downstream VLA performance, demonstrating superior time efficiency for both training and inference.

\begin{figure}[t]
  \centering
  \includegraphics[width=\linewidth]{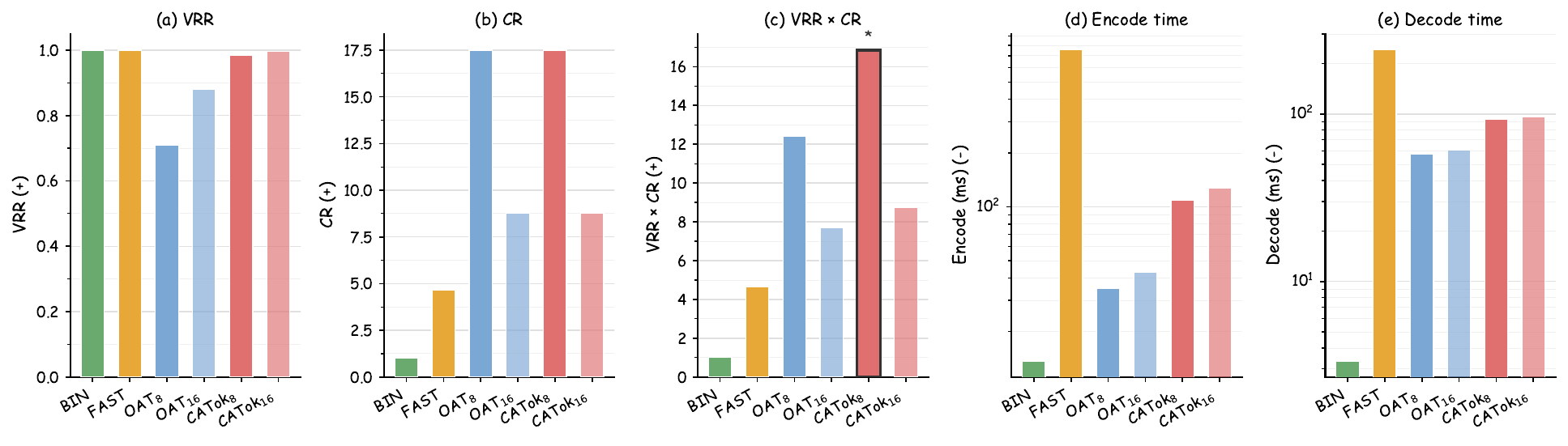}
  \caption{
  \textbf{Intrinsic properties of action tokenizers.}
  \textbf{(a--c) Fidelity vs.\ compression efficiency.}
  VRR measures reconstruction quality; CR measures compression ratio; VRR$\times$CR jointly captures the fidelity--compression tradeoff.
  \modelname$_8$ achieves the highest VRR$\times$CR ($16.85$), outperforming all baselines at the same compression level.
  \textbf{(d--e) Inference efficiency.}
  Encode and decode latencies are shown on a log scale.
  The subscript $k$ in OAT$_k$ and \modelname$_k$ denotes the number of tokens.
  }
  \label{fig:tokenizer_properties}
\end{figure}

\subsection{Ablations and Analysis}
We conduct ablation studies on key components and design choices in \modelname. Specifically, we analyze the effect of the flow-matching head, where the results show that the complete design consistently achieves the best performance. In addition, we study the impact of the number of tokens, a key design choice in \modelname, on downstream VLA performance. Detailed ablation settings and additional results are provided in Appendix~\ref{appendix:ablation}.

\section{Conclusion}

We introduced \modelname{}, a causal action tokenizer for purely autoregressive VLA models.
By transferring the stage-wise causal structure of flow matching into the token space through conditional annealing, \modelname{} learns compact action tokens with causal ordering and generative semantics.
The resulting token sequence enables autoregressive policies to generate continuous action chunks through a coarse-to-fine factorization, while a frozen MMDiT-based flow-matching decoder preserves high-fidelity action reconstruction.
Experiments across multiple simulation benchmarks and real-world robotic manipulation tasks demonstrate that \modelname{} improves downstream task performance, training efficiency, and the fidelity--compression tradeoff over existing action tokenizers, while maintaining competitive inference efficiency.
Our analysis further verifies the causal ordering and generative semantics of the learned token sequence.
These results highlight causal and generative action tokenization as a promising approach for scalable purely autoregressive VLA systems.

\textbf{Limitations.}
Despite the promising results, our real-world evaluation is currently limited to a single robot platform and a finite set of manipulation tasks.
Therefore, the robustness of \modelname{} across diverse robot embodiments, sensing configurations, environmental conditions, and dynamics remains unexplored.
In addition, our experiments are conducted on relatively constrained datasets, and we have not yet investigated the scalability of \modelname{} on large-scale and diverse cross-embodiment robotic data.
Future work will evaluate \modelname{} across broader real-world embodiments and large-scale robotic datasets to study its scalability and generalization under more diverse real-world settings.

\newpage

\bibliographystyle{unsrt}
\bibliography{ref/refs}

\newpage
\appendix
\section{Implementation Details}

\newcolumntype{Y}{>{\raggedright\arraybackslash}X}
\newcommand{\tbd}{\textsc{TBD}}
\label{appendix:implementation}

Table~\ref{tab:implementation_details} provides implementation details for the action tokenizer training, VLA training, and evaluation protocols used in our experiments. In particular, for LIBERO and Bridge, we train a tokenizer on the combined datasets with mix ratios of 1.0 and 5.0, respectively, and then train the VLA separately on each dataset. For RoboTwin, we train both the tokenizer and the VLA on the full RoboTwin dataset.

\begin{table*}[htb]
\centering
\caption{Benchmark-specific implementation details.}
\label{tab:implementation_details}
\small
\setlength{\tabcolsep}{5pt}
\renewcommand{\arraystretch}{1.12}
\begin{tabularx}{\textwidth}{@{}lYYY@{}}
\toprule
\textbf{Setting} 
& \textbf{LIBERO} 
& \textbf{Simpler-Bridge} 
& \textbf{RoboTwin 2.0} \\
\midrule

\multicolumn{4}{@{}l}{\textit{Benchmark and embodiment}} \\
\midrule
Embodiment 
& Franka 
& WidowX 
& Bimanual ALOHA \\

Action DoF 
& 7
& 7
& 14 \\

Training data 
& Official LIBERO dataset 
& Bridge dataset 
& Clean and randomized RoboTwin 2.0 trajectories \\

Evaluation protocol 
& \makecell[l]{4 suites\\10 tasks per suite\\50 rollouts per task}
& \makecell[l]{4 WidowX tasks\\in SimplerEnv\\24 rollouts per task}
& \makecell[l]{50 tasks\\20 rollouts per task} \\

\addlinespace[0.5em]
\midrule
\multicolumn{4}{@{}l}{\textit{Action tokenizer configuration}} \\
\midrule
Action horizon $H$ 
& 20 
& 10 
& 20 \\

Number of action tokens $K$ 
& 16
& 16 
& 32 \\

Codebook size $|\mathcal{C}|$ 
& 4096 
& 4096 
& 4096 \\

Tokenizer training steps 
& 300 000
& 300 000
& 200 000 \\

Tokenizer batch size 
& 512 
& 512 
& 512 \\

Tokenizer learning rate 
& $1 \times 10^{-4}$ 
& $1 \times 10^{-4}$ 
& $1 \times 10^{-4}$  \\

Tokenizer weight decay 
& $1 \times 10^{-2}$ 
& $1 \times 10^{-2}$ 
& $1 \times 10^{-2}$  \\

Tokenizer warmup steps
& 1000
& 1000
& 1000 \\

\addlinespace[0.5em]
\midrule
\multicolumn{4}{@{}l}{\textit{VLA training configuration}} \\
\midrule
VLA backbone 
& $\pi_0$-FAST-base 
& Qwen3-VL-4B-Instruct 
& Qwen3-VL-4B-Instruct \\

VLA training steps
& 30 000
& 60 000
& 40 000 \\

VLA batch size 
& 32 
& 64 
& 32 \\

VLA learning rate 
& $2.5 \times 10^{-5}$  
& $1 \times 10^{-5}$  
& $1 \times 10^{-5}$ \\

VLA weight decay
& $1 \times 10^{-10}$
& $1 \times 10^{-8}$
& $1 \times 10^{-8}$ \\

Number of GPUs 
& 4$\times$\text{H100 GPUs}
& 4$\times$\text{H100 GPUs}
& 4$\times$\text{H100 GPUs} \\

\bottomrule
\end{tabularx}
\end{table*}

\section{Evaluation Metrics}
\label{appendix:metrics}
In this section, we provide the detailed definitions and computation protocols for the evaluation metrics used to assess the intrinsic properties and efficiency of action tokenizers.
\subsection{Valid Reconstruction Rate (VRR)}
To evaluate reconstruction fidelity, we adopt the Valid Reconstruction Rate (VRR)~\cite{liu2025faster}, defined as the proportion of reconstructed action chunks whose normalized reconstruction error falls below a predefined threshold:
\begin{align}
    \mathrm{VRR}
    =
    \frac{N_{\mathrm{valid}}}{N_{\mathrm{total}}},
    \quad
    N_{\mathrm{valid}}
    =
    \sum_{i=1}^{N_{\mathrm{total}}}
    \mathbf{1}
    \left(
    \frac{
    \|A_i^{\mathrm{pred}}-A_i^{\mathrm{gt}}\|_2
    }{H}
    < \sigma
    \right),
\end{align}
where $A_i^{\mathrm{pred}}, A_i^{\mathrm{gt}} \in \mathbb{R}^{H \times D}$ denote the reconstructed and ground-truth action chunks for the $i$-th sample, respectively. Here, $H$ is the prediction horizon and $D$ is the action dimension. The $\ell_2$ norm is computed over both temporal and action dimensions, and normalization by $H$ yields the average per-step reconstruction error. $\sigma$ denotes a predefined threshold, and $\mathbf{1}(\cdot)$ is the indicator function.

\subsection{Compression Rate (CR)}
To measure reconstruction efficiency, we use the Compression Rate (CR), defined as the ratio between the size of the original continuous action representation $N_{\text{raw}}$ and the compressed discrete token sequence $N_{\text{token}}$:
\begin{align}
    \mathrm{CR}
    =
    \frac{N_{\mathrm{raw}}}{N_{\mathrm{token}}},
    \quad
    N_{\mathrm{raw}}
    =
    B \times H \times D,
\end{align}
where $B$ denotes the batch size, $H$ is the action horizon, and $D$ is the action dimension. A higher CR indicates more compact action representations.


\subsection{Encoding and Decoding Latency}
To evaluate the efficiency of action tokenization during both training and inference, we measure the average encoding and decoding latency. Since action encoding is performed during training, encoding latency directly affects training efficiency. In contrast, decoding latency primarily impacts inference-time efficiency. Specifically:
\begin{itemize}
    \item \textbf{Encoding time (\text{encode\_time})}: the average time required to convert continuous action chunks into discrete token sequences.
    \item \textbf{Decoding time (\text{decode\_time})}: the average time required to reconstruct continuous actions from discrete tokens.
\end{itemize}

All latency measurements are averaged across batches under the same hardware and evaluation settings.

\section{Ablations and Analysis}
\label{appendix:ablation}
In this section, we provide detailed ablation settings, additional experimental results, and further analysis for the key design components and design choices in \modelname.

\paragraph{Conditional Annealing and Architecture Ablations.}
To isolate the contribution of conditional annealing from the effects of decoder architecture and model capacity, we consider four variants: (A) our full model with conditional annealing, (B) \modelname~without conditional annealing, (C) a VQ-VAE with the same MMDiT decoder and comparable parameter count trained with random token masking, and (D) a VQ-VAE with the same MMDiT decoder and comparable parameter count without random masking.

\begin{table}[t]
\centering
\small
\setlength{\tabcolsep}{4.5pt}
\renewcommand{\arraystretch}{1.1}
\caption{Controlled ablation of conditional annealing, random masking, and decoder architecture.}
\label{tab:annealing_ablation}
\begin{tabular}{lccccc}
\toprule
Method & Spatial & Object & Goal & Long & Avg. \\
\midrule
A - Ours & 0.978 & 0.994 & 0.954 & 0.910 & 0.959 \\
B - Ours w/o conditional annealing & 0.920 & 0.980 & 0.918 & 0.834 & 0.914 \\
C - VQ-VAE + MMDiT + random masking & 0.724 & 0.924 & 0.756 & 0.542 & 0.737 \\
D - VQ-VAE + MMDiT & 0.614 & 0.796 & 0.296 & 0.240 & 0.486 \\
\bottomrule
\end{tabular}
\end{table}

Removing conditional annealing (A $\rightarrow$ B) leads to a clear performance drop from 0.959 to 0.914, demonstrating its importance beyond the underlying flow-matching decoder and transformer architecture. Moreover, both C and D remain substantially below our method despite using the same MMDiT decoder and comparable model capacity, showing that the improvement cannot be explained solely by decoder architecture or model capacity. Notably, C already employs random token masking yet achieves only 0.737, indicating that random masking alone is insufficient to induce the desired causal structure in the token space. Together, these controlled comparisons demonstrate that conditional annealing is a critical factor in organizing token representations into a structured and causally ordered representation.

\begin{figure}[t]
    \centering
    \includegraphics[width=\linewidth]{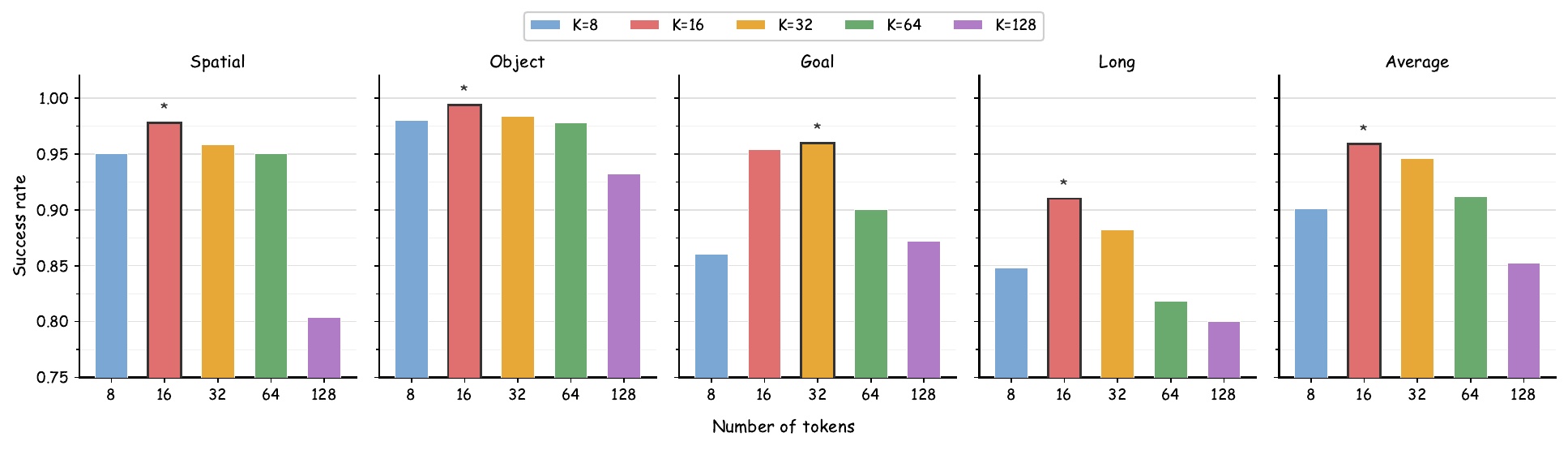}
    \caption{Effect of the number of action tokens on downstream VLA performance on the LIBERO benchmark. Performance improves from $8$ to $16$ tokens but degrades with larger token budgets, with $16$ tokens achieving the best overall performance.}
    \label{fig:num-tokens}
\end{figure}

\paragraph{Number of Tokens.}
We further analyze the impact of the number of action tokens in \modelname~ on downstream VLA performance. Specifically, we evaluate \modelname~ with different token budgets ($8, 16, 32, 64,$ and $128$ tokens) while keeping all other training and architecture settings fixed. The results on the LIBERO benchmark are shown in Figure~\ref{fig:num-tokens}. Increasing the number of tokens from $8$ to $16$ consistently improves downstream VLA performance, while further increasing the token budget beyond $16$ leads to performance degradation. We attribute this degradation to the increased difficulty of autoregressive modeling over excessively long token sequences, which reduces token efficiency and negatively impacts downstream policy learning. Overall, $16$ tokens provide the best balance between representation capacity and autoregressive learning efficiency.

\subsection{Causality Validation}
\label{appendix:causality-validation}

To further verify that the learned token ordering corresponds to genuine
causal roles in action generation, we perform intervention experiments using
the frozen native decoder of \modelname{}.
Specifically, we consider two types of interventions: \textit{donor-swap}
and \textit{single-token removal}.
Both experiments directly modify individual token positions while keeping
the remaining tokens unchanged, allowing us to measure the causal influence
of each token on the reconstructed action trajectory.

\paragraph{Donor-Swap Intervention.}
We first evaluate the influence of each token position by replacing the
token at position $k$ with the corresponding token from another action
chunk, while keeping all other tokens unchanged.
We then decode the intervened token sequence using the native
flow-matching decoder and measure the resulting deviation from the
original reconstruction.
By performing this intervention at different token positions and
flow-matching time steps, we obtain a position-dependent influence map.
If the tokens follow a causal coarse-to-fine ordering, early tokens should
primarily affect coarse action structure, while later tokens should exert
stronger influence during later stages of generation.

\paragraph{Single-Token Removal.}
We further perform a single-token removal experiment by replacing the token
at position $k$ with a neutral embedding while keeping all other tokens
unchanged.
The resulting action reconstruction is compared with the reconstruction
from the complete token sequence.
This experiment isolates the contribution of individual token positions
and provides a complementary measure of their generative influence.

\begin{figure}[ht]
  \centering
  \begin{subfigure}{\textwidth}
    \centering
    \includegraphics[width=\textwidth]{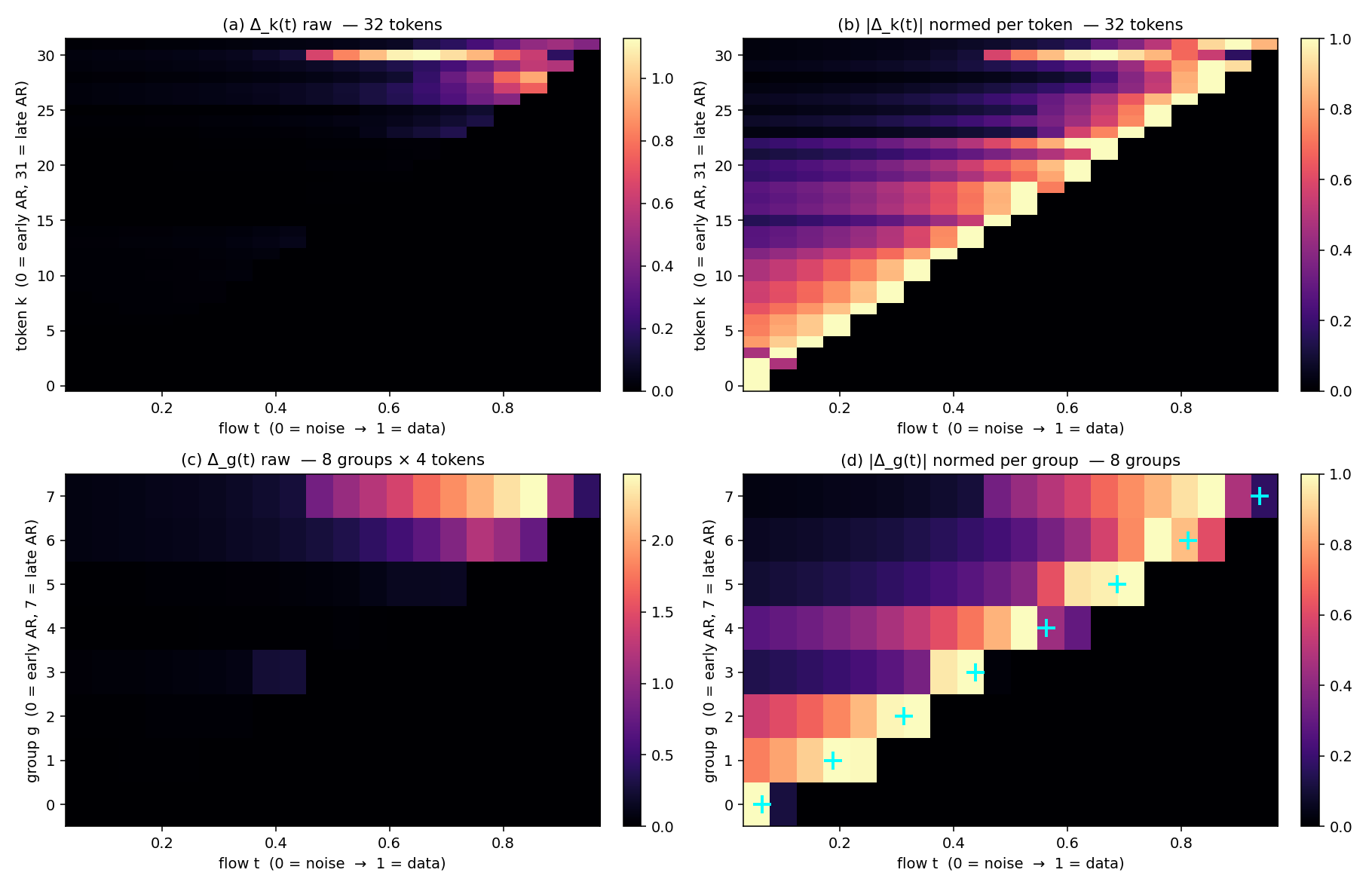}
    \caption{Donor-swap influence matrix $\Delta(t)$. Replacing token $k$ with a donor's same-position embedding yields a position- and flow-stage-dependent influence map with triangular support, confirming the conditional-annealing schedule.}
    \label{subfig:donor_swap}
  \end{subfigure}

  \par\medskip

  \begin{subfigure}{0.85\textwidth}
    \centering
    \includegraphics[width=\linewidth]{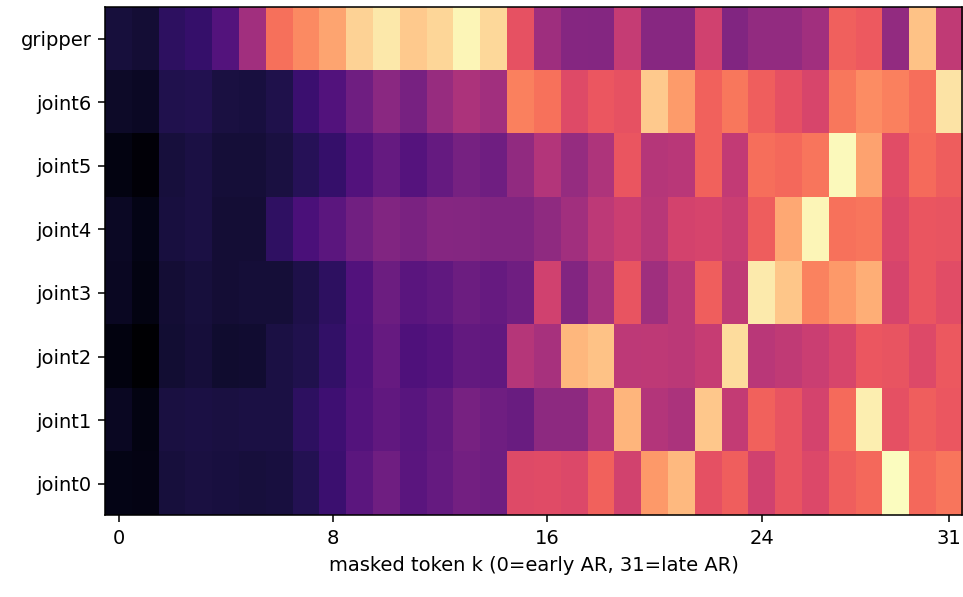}
    \caption{Single-token removal. Zeroing token $k$ and decoding the intervened sequence isolates each position's generative contribution; the per-dimension footprint progresses from coarse gripper/task-state information (early tokens) to fine joint/trajectory-detail information (late tokens).}
    \label{subfig:single_token_removal}
  \end{subfigure}

  \caption{Causality validation via native-decoder interventions. Both donor-swap (a) and single-token removal (b) exhibit a clear position-dependent structure: early tokens have stronger influence on coarse action components and earlier generation stages, whereas later tokens increasingly affect fine-grained action details, confirming that the learned token ordering corresponds to meaningful causal roles rather than an arbitrary positional ordering.}
  \label{fig:causality-validation}
\end{figure}

\paragraph{Results.}
As shown in Figure~\ref{fig:causality-validation}, the intervention effects
exhibit a clear position-dependent structure.
Early tokens have stronger influence on coarse action components and earlier
generation stages, whereas later tokens increasingly affect fine-grained
action details.
The consistent stage-specific influence observed under both donor-swap and
single-token removal interventions provides direct evidence that the token
ordering learned by \modelname{} corresponds to meaningful causal roles
rather than merely reflecting an arbitrary positional ordering.

\section{Full Robotwin 2.0 Benchmark}
Table~\ref{tab:robotwin_per_task} provides the full per-task RoboTwin 2.0 results corresponding to the aggregate scores in Table~\ref{tab:success-rate}. 
We report success rates for both clean and randomized evaluation settings across all 50 tasks, with 20 rollouts per task.
\label{appendix:Robotwin}
\begin{table*}[t]
\centering
\caption{Per-task success rates on RoboTwin 2.0.}
\label{tab:robotwin_per_task}
\small
\setlength{\tabcolsep}{4.1pt}
\renewcommand{\arraystretch}{1.0}
\resizebox{0.95\textwidth}{!}{
\begin{tabular}{@{}lcccccccc@{}}
\toprule
\multirow{2}{*}{\textbf{Simulation Task}}
& \multicolumn{2}{c}{\textbf{BIN}}
& \multicolumn{2}{c}{\textbf{FAST}}
& \multicolumn{2}{c}{\textbf{OAT}}
& \multicolumn{2}{c}{\textbf{\modelname}} \\
\cmidrule(lr){2-3}
\cmidrule(lr){4-5}
\cmidrule(lr){6-7}
\cmidrule(lr){8-9}
& Clean & Rand. & Clean & Rand. & Clean & Rand. & Clean & Rand. \\
\midrule
\textit{Adjust Bottle}             & 80\% & 95\% & 90\% & 100\% & 100\% & 100\% & 90\% & 100\% \\
\textit{Beat Block Hammer}         & 15\% & 20\% & 25\% & 50\% & 15\% & 5\% & 65\% & 65\% \\
\textit{Blocks Ranking RGB}        & 0\% & 0\% & 10\% & 10\% & 5\% & 0\% & 5\% & 0\% \\
\textit{Blocks Ranking Size}       & 0\% & 0\% & 0\% & 5\% & 5\% & 0\% & 5\% & 10\% \\
\textit{Click Alarmclock}          & 65\% & 60\% & 90\% & 75\% & 60\% & 70\% & 65\% & 80\% \\
\textit{Click Bell}                & 95\% & 85\% & 80\% & 80\% & 75\% & 50\% & 95\% & 95\% \\
\textit{Dump Bin Bigbin}           & 30\% & 30\% & 95\% & 75\% & 15\% & 20\% & 65\% & 70\% \\
\textit{Grab Roller}               & 65\% & 50\% & 100\% & 100\% & 55\% & 60\% & 100\% & 100\% \\
\textit{Handover Block}            & 0\% & 0\% & 0\% & 0\% & 0\% & 0\% & 10\% & 10\% \\
\textit{Handover Mic}              & 10\% & 10\% & 55\% & 50\% & 5\% & 10\% & 60\% & 65\% \\
\textit{Hanging Mug}               & 0\% & 0\% & 10\% & 15\% & 0\% & 0\% & 15\% & 5\% \\
\textit{Lift Pot}                  & 25\% & 0\% & 45\% & 45\% & 0\% & 0\% & 30\% & 40\% \\
\textit{Move Can Pot}              & 0\% & 0\% & 25\% & 50\% & 15\% & 15\% & 50\% & 55\% \\
\textit{Move Pillbottle Pad}       & 0\% & 10\% & 55\% & 40\% & 15\% & 25\% & 50\% & 70\% \\
\textit{Move Playingcard Away}     & 30\% & 65\% & 65\% & 80\% & 45\% & 60\% & 90\% & 85\% \\
\textit{Move Stapler Pad}          & 0\% & 5\% & 10\% & 10\% & 5\% & 0\% & 15\% & 10\% \\
\textit{Open Laptop}               & 0\% & 15\% & 70\% & 55\% & 20\% & 20\% & 75\% & 90\% \\
\textit{Open Microwave}            & 0\% & 5\% & 15\% & 20\% & 20\% & 15\% & 20\% & 35\% \\
\textit{Pick Diverse Bottles}      & 0\% & 0\% & 40\% & 45\% & 0\% & 5\% & 35\% & 40\% \\
\textit{Pick Dual Bottles}         & 0\% & 0\% & 30\% & 55\% & 0\% & 5\% & 60\% & 40\% \\
\textit{Place A2B Left}            & 40\% & 30\% & 85\% & 80\% & 40\% & 40\% & 75\% & 80\% \\
\textit{Place A2B Right}           & 35\% & 25\% & 65\% & 70\% & 45\% & 35\% & 65\% & 70\% \\
\textit{Place Bread Basket}        & 25\% & 20\% & 70\% & 40\% & 20\% & 20\% & 45\% & 65\% \\
\textit{Place Bread Skillet}       & 40\% & 35\% & 50\% & 50\% & 15\% & 10\% & 45\% & 60\% \\
\textit{Place Burger Fries}        & 5\% & 10\% & 80\% & 70\% & 25\% & 40\% & 80\% & 80\% \\
\textit{Place Can Basket}          & 0\% & 5\% & 20\% & 30\% & 0\% & 5\% & 30\% & 50\% \\
\textit{Place Cans Plasticbox}     & 0\% & 0\% & 50\% & 25\% & 0\% & 0\% & 30\% & 35\% \\
\textit{Place Container Plate}     & 55\% & 60\% & 85\% & 96\% & 65\% & 50\% & 90\% & 90\% \\
\textit{Place Dual Shoes}          & 5\% & 0\% & 10\% & 20\% & 0\% & 0\% & 0\% & 15\% \\
\textit{Place Empty Cup}           & 35\% & 35\% & 80\% & 85\% & 35\% & 10\% & 50\% & 75\% \\
\textit{Place Fan}                 & 0\% & 10\% & 40\% & 40\% & 5\% & 5\% & 25\% & 30\% \\
\textit{Place Mouse Pad}           & 5\% & 10\% & 30\% & 30\% & 5\% & 5\% & 30\% & 55\% \\
\textit{Place Object Basket}       & 5\% & 5\% & 45\% & 25\% & 20\% & 25\% & 50\% & 60\% \\
\textit{Place Object Scale}        & 10\% & 30\% & 70\% & 45\% & 20\% & 30\% & 40\% & 50\% \\
\textit{Place Object Stand}        & 35\% & 45\% & 80\% & 90\% & 35\% & 50\% & 75\% & 85\% \\
\textit{Place Phone Stand}         & 25\% & 0\% & 65\% & 55\% & 10\% & 0\% & 40\% & 75\% \\
\textit{Place Shoe}                & 25\% & 25\% & 50\% & 40\% & 25\% & 50\% & 75\% & 50\% \\
\textit{Press Stapler}             & 45\% & 45\% & 60\% & 55\% & 55\% & 75\% & 85\% & 85\% \\
\textit{Put Bottles Dustbin}       & 0\% & 0\% & 5\% & 0\% & 0\% & 0\% & 5\% & 0\% \\
\textit{Put Object Cabinet}        & 0\% & 0\% & 25\% & 35\% & 0\% & 0\% & 45\% & 40\% \\
\textit{Rotate QRcode}             & 10\% & 10\% & 30\% & 35\% & 0\% & 5\% & 60\% & 35\% \\
\textit{Scan Object}               & 0\% & 10\% & 20\% & 25\% & 0\% & 5\% & 25\% & 20\% \\
\textit{Shake Bottle}              & 95\% & 85\% & 100\% & 100\% & 85\% & 100\% & 90\% & 95\% \\
\textit{Shake Bottle Horizontally} & 90\% & 95\% & 100\% & 95\% & 90\% & 85\% & 100\% & 85\% \\
\textit{Stack Blocks Three}        & 0\% & 0\% & 0\% & 10\% & 0\% & 0\% & 0\% & 0\% \\
\textit{Stack Blocks Two}          & 10\% & 10\% & 50\% & 40\% & 15\% & 5\% & 40\% & 50\% \\
\textit{Stack Bowls Three}         & 0\% & 0\% & 10\% & 10\% & 0\% & 0\% & 20\% & 15\% \\
\textit{Stack Bowls Two}           & 25\% & 25\% & 40\% & 75\% & 45\% & 20\% & 50\% & 60\% \\
\textit{Stamp Seal}                & 5\% & 15\% & 25\% & 40\% & 15\% & 20\% & 40\% & 40\% \\
\textit{Turn Switch}               & 25\% & 15\% & 40\% & 20\% & 20\% & 15\% & 40\% & 40\% \\
\midrule
\textbf{Average}                   & 21.3\% & 22.1\% & 47.8\% & 47.82\% & 22.9\% & 23.3\% & 48.9\% & 53.1\% \\
\bottomrule
\end{tabular}
}
\end{table*}

\end{document}